\documentclass{article}
\usepackage{spconf}
\usepackage{amsfonts}
\usepackage{amsmath}
\usepackage{graphicx}

\usepackage[T2A]{fontenc}

\usepackage[utf8]{inputenc}
\usepackage[russian, english]{babel}

\usepackage{latexsym}

\usepackage{url}

\title{Guess who?\\
\small{Multilingual approach for the automated generation of author-stylized poetry}}

\twoauthors
  {Alexey Tikhonov}
	{Yandex\\
	Karl-Liebknecht strasse 1, Berlin \\
  {\tt altsoph@gmail.com}}
  {Ivan P. Yamshchikov}
	{Max Planck Institute for Mathematics in the Sciences\\
	Inselstrasse 22, Leipzig\\
	{\tt ivan@yamshchikov.info}}

\begin{document}
%
\maketitle
\begin{abstract}
 This paper addresses the problem of stylized text generation in a multilingual setup. A version of a language model based on a long short-term memory (LSTM) artificial neural network with extended phonetic and semantic embeddings is used for stylized poetry generation. The quality of the resulting poems generated by the network is estimated through bilingual evaluation understudy (BLEU), a survey and a new cross-entropy based metric that is suggested for the problems of such type. The experiments show that the proposed model consistently outperforms random sample and vanilla-LSTM baselines, humans also tend to associate machine generated texts with the target author. 
\end{abstract}

\section{Introduction}

The problem of making machine-generated text feel more authentic has a number of industrial and scientific applications, see, for example, \cite{Livingstone} or \cite{Dix}. Most modern generative models are trained on huge corpora of texts which include different contributions from various authors. It is no surprise that texts produced with such models are often not perceived as natural and are characterized as flat and non-human since humans have recognizable writing and communication styles. One of the possible ways to approach this problem is to propose a model that would generate texts resembling the style of a particular author within the training data set. In this paper we quantify this stylistic similarity, propose a generative model that captures it, and show that it outperforms a standard long short-term memory (LSTM) model used for text generation. We strongly believe that the proposed model is also applicable to prose or dialogue setup, but we carry out our experiments using poetry for a number of reasons. First of all, it is harder to train a model on poetic texts since the absolute size of the training corpus for poetry would be inevitably smaller than a corpus for prose which would include a comparable number of authors.  On the other hand from a stylistic perspective, poetry is often believed to be more expressive than prose, so one can better see if the generated output is indeed stylized. This factor significantly affects any kind of qualitative tests that involve subjective human judgement.

The contribution of this paper is four-fold: (1) we formalize the problem of stylized poetry generation; (2) we suggest a {\em sample cross-entropy} metric to measure the quality of author stylization; (3) we propose an LSTM with extended phonetic and semantic embeddings and quantify the quality of the obtained stylized poems both subjectively through a survey and objectively with sample cross-entropy and BLEU metrics; (4) we demonstrate that the proposed approach works in a multilingual setting, providing examples in English and in Russian. 

\section{Related work}

The idea that computers can generate poetry algorithmically dates back more than half a century, see \cite{Wheatley}. A detailed taxonomy of generative poetry techniques can be found in \cite{Lamb}. In this paper, we specifically focus on RNN-based generative models, so let us briefly mention several contributions relevant to the further discussion.  

Recently  \cite{Lipton}, \cite{Kiddon}, \cite{Lebret}, \cite{Radford}, \cite{Tang}, \cite{Hu} have developed RNN-based generative or generative adversarial models for controlled text generation that were focused on the {\em content} and {\em semantics} of the output, yet did not tale the stylistic aspects of the generated texts into consideration.  In \cite{Li2016APN} the authors came up with a persona-based models for handling the issue of speaker consistency in neural response generation. They focused on the speaker consistency in the dialogue setup and demonstrated that the model could show better results than baseline sequence-to-sequence models. In \cite{Sutskever} the authors demonstrated that a character-based recurrent neural network with gated connections can successfully generate texts that resemble news or Wikipedia articles. In \cite{Graves} it was shown that comparable prosaic texts can be generated with LSTM networks as well. There are a number of works specifically focused on Chinese classical poetry generation, for example \cite{Hezhou}, \cite{Yan1}, \cite{Yan2}, \cite{Yi} or \cite{ZhangJ}, however interesting contributions in the area of generative poetry in languages other than Chinese or in a multilingual setting are relatively rare. One could mention the paper by \cite{Ghazvininejad} where an algorithm generates a poem in line with a topic given by the user and the paper by \cite{Potash} in which the authors generate stylized rap lyrics with LSTM trained on a rap poetry corpus.

A literary style is actually not an obvious notion. There is a number of style transfer papers that deal with different aspects of literary styles. These could be a sentiment of a text (see \cite{Shen} or \cite{li}), it's politeness \cite{Sennrich} or a so-called style of the time (see \cite{Hughes}). The style of the time aspect is specifically addressed by \cite{Jhamtani} and by \cite{Carlson}. A paper by \cite{Fu} generalizes these ideas measuring the success of a particular style aspect with a specifically trained classifier. However, the problem of style transfer differs from the stylized text generation significantly since as it was shown in \cite{guu} an existent human-written source used to control the saliency of the output can significantly improve the quality of the resulting texts. The generative model does not have such input and generates stylized texts from scratch, in this sense our problem set-up is similar to \cite{Ficler}, but differs in the area of application and the definition of style. Specifically, we believe that style of the text should be implicitly defined by the corpus rather than be a set of binary, human-defined characteristics \cite{wrong}.

\section{Generation of stylized texts}\label{formulation}

Let us consider a corpus $C = \{ T_i \}^{M}_{i = 0}$ of $M$ literary texts written in one natural language. Every text of length $l$ is a sequence $T_i = (w_j)^{l}_{j = 0}$ where words (denoted here as $w_j$) are drawn from a vocabulary set $V = \{ w_j \}^{L}_{j=1}$, where $L$ is the size of a given vocabulary.

In a generative context, the standard language model predicts the next word $w_k$ using a conditional probability $P(w_k | (w_i)^{k-1}_{i=0})$. Neural networks have been widely considered as the most promising technique for language modeling since \cite{Bengio}, see also \cite{Morin} and \cite{Mnih}. One of the key advantages of neural networks is that they help to avoid the dimensionality curse  \cite{Mikolov} of a classical language model obtaining an effective mapping $Y : (C, \mathbb{R}^{m}, F) \rightarrow \mathbb{R}^{d}$ and then train a model such that $G(C) : \mathbb{R}^{d}  \rightarrow \{T^{G}_i \}$.

In the majority of works on text generation, one uses additional observable information to improve the general performance of the model \cite{Shi}. That is, if authors define a certain performance metric $D$ (such as BLEU, F1, etc.) one usually tries to minimize $D(\{ T_i \},  \{T^{G}_i \})$, where $\{ T_i \}$ is usually a randomized sample of $C$. We on the other hand suggest to look for a {\em stylization model} $G(C|S)$ that takes into consideration a subset $S$ of continuos and categorial variables out of $(\mathbb{R}^{m}, F)$ and a metric $D$  so that
\begin{equation}
G(C|S): 
\begin{cases} \label{problem}
	( C, \mathbb{R}^{m}, F) \rightarrow \{T^{G}_i \} \\
	\{ T^{G}_i | S \} \sim \{ T_i | S \} \hspace{2pt} \text{w.r.t.} \hspace{2pt} D 
\end{cases}
\end{equation}
A distinct difference in this approach is that we train our model on all information available to us, i.e. $( C, \mathbb{R}^{m}, F)$, and yet we are not interested in its overall performance, but rather test it on a certain domain $S$. The motivation here is in some sense similar to one-shot learning, see \cite{fei06} and, generally, transfer learning, see \cite{pratt12}, and author-attribution method, see \cite{Bagnall}. A model uses information on the structure of the broader domain of data. Such information is formally exogenous to the problem in its' narrow formulation, but it can improve the performance of the model. Stylization model has a number of interesting benefits in contrast to a language model. First of all, it naturally implies customization. If we want to control certain parameters of the model, we include them in $S$ and can expect that output $\{ T^{G}_i | S \}$ will resemble original texts $ \{ T_i | S \}$ that satisfy $S$ conditions. This makes such an approach easily applicable to, say, personalized interfaces. On the other hand, one would expect that due to its umbrella structure in which $G(C|S)$ learns from the whole corpus $(C, \mathbb{R}^{m}, F)$ such a model would outperform a set of smaller models obtained from different subsamples of $C$. Artificial neural networks are known to generalize very well, which lets one speculate that system that is trained on the whole corpus $C$ would be generally outperforming the system that uses less information for training. 

Further in this paper, we describe an artificial neural network that uses the name of an author of a poetic text as a condition $S$. We show that this model can generate lyrics that resemble the text written by a given author both objectively (in terms of sample cross-entropy that we define further and BLEU) and subjectively (based on a survey of respondents). This model has been trained with English and Russian, and we do not see obstacles for its application to the corpora in other languages.

\section{Model}

\begin{figure}[!ht]\centering 
\includegraphics[width=\linewidth]{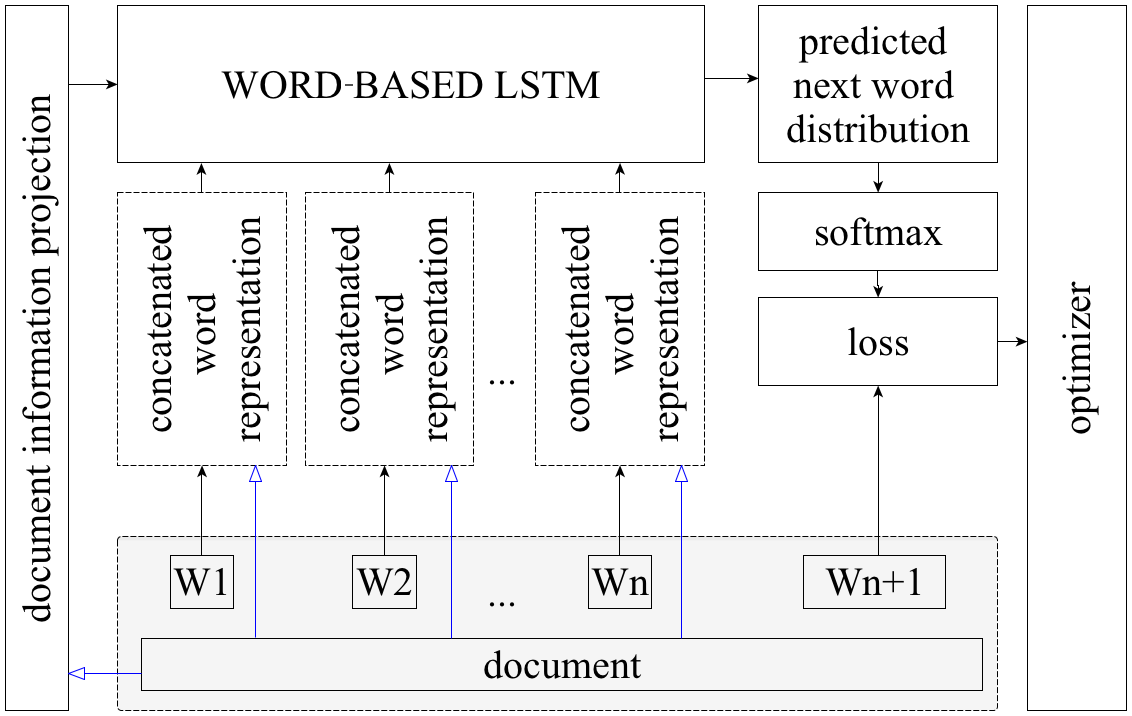} 
\caption{The scheme of the language model used. Document information projections are highlighted with blue and white arrows. The projections on a state space of the corresponding dimension is achieved with simple matrix multiplication of document embeddings.}
\label{fig:mod}
\end{figure}

We use an LSTM-based language model that predicts the $w_{n+1}$ word based on $w_1, ... , w_n$ previous inputs and some other parameters of the modeled sequence. One of the most widespread approaches for passing the needed parameter to the network is to write it in its initial state.  A general weakness of this approach is that the network 'forgets' the general parameters of the document as the generated sequence gets longer. Since we want to develop a model in line with the formulation given in (\ref{problem}) we support our model at every step with the embeddings of the document that is currently being analyzed. This idea differentiates our approach from a classical word-based LSTM and was, for example, used in \cite{TiYa} to facilitate stylized music generation. A schematic picture of the model is shown in Figure \ref{fig:mod}, document information projections are highlighted with blue and white arrows. We used an LSTM with 1152-dimensional input and 512-dimensional state.  

\begin{figure}[!ht]\centering 
\includegraphics[width=\linewidth]{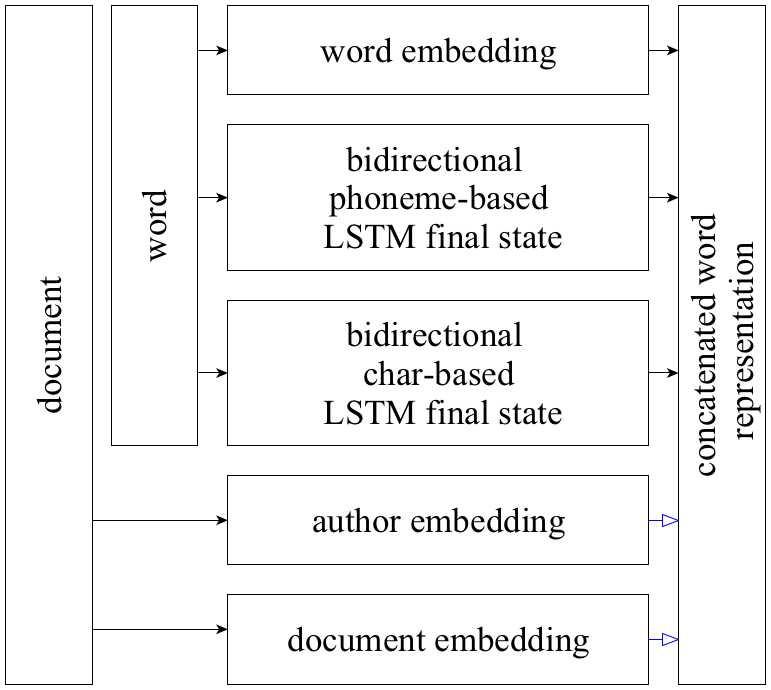} 
\caption{Concatenated word representation.}
\label{fig:emb}
\end{figure}

Another key feature of the proposed model is a concatenated word representation shown schematically in Figure \ref{fig:emb}. Information about the document (512-dimensional projection of a concatenated author and document embeddings) is included at every step. Final states of two char bidirectional LSTMs with a 128-dimensional vector are also concatenated into a word embedding. One of the LSTMs works with letters from a char-representation of the word whereas another uses phonemes of the International Phonetic Alphabet\footnotemark, employing an heuristics to transcribe words into phonemes. A somewhat similar idea, but with convolutional neural networks rather than with LSTMs, was proposed in \cite{Jozefowicz}, but the bidirectional LSTM approach is new to our knowledge.

\addtocounter{footnote}{0}
\footnotetext{\noindent{https://en.wikipedia.org/wiki/International \_Phonetic\_Alphabet}}

In the Section \ref{ex} we describe a series of objective and subjective tests that we ran across a generated output $ \{ T_i | S \}$, but first let us briefly describe the datasets used for training.

\section{Datasets}

We have trained our model on two datasets of English and Russian poetry. The datasets were proprietary ones and were already available. All punctuation was deleted, every character was transferred to a lower case. No other preprocessing was made. The datasets sizes can be found in Table \ref{tab:dt}.

\begin{table}[t!]
\small
\begin{tabular}{lllll}
\bf  & \bf N. of & \bf Size of  & \bf N. of  & \bf Size \\ 
\bf  & \bf documents & \bf vocab.  & \bf authors & \bf   \\ \hline
 English & 110000  & 165000  & 19000 & 150 Mb \\
Russian &  330000  & 400000  & 1700 & 140 Mb \\
\end{tabular}
\caption{\label{tab:dt} Parameters of the training datasets.}
\end{table}

During the training phase we tokenize the beginning and ending of every text $T_i$, so that in the generation phase the network is initialized with a special 'start' token and is conditioned on values of document parameters $S$. In this paper we test the proposed mechanism for the stylized text generation with one categorical variable - the name of the author. We trained the model for English (running tests on lyrics of William Shakespeare, Edgar Allan Poe, Lewis Carroll, Oscar Wilde and Bob Marley as well as lyrics of the American band Nirvana and UK band Muse) and Russian (Alexander Pushkin, Sergey Esenin, Joseph Brodsky, Egor Letov and Zemfira Ramazanova). As one can see in Table \ref{tab:dt}, there were far more authors in the dataset, but we chose more prominent ones who are known for their poetic styles and therefore could be more readily identified by an educated reader who is fluent in the target language. We want to emphasize that we do not see any excessive difficulties in implementation of the proposed model for other languages for which one can form a training corpus $C$ and provide a phonetically transcribed vocabulary $V_p$. 

Table \ref{tab:ex} shows some generated stylized poetry examples. The model captures syntactic characteristics of the author (note the double negation in the first and the last line of generated Marley) alongside with the vocabulary ('burden', 'darkness', 'fears' could be subjectively associated with gothic lyrics of Poe, whereas 'sunshine', 'fun', 'fighting every rule' could be associated with positive yet rebellious reggae music). 

\begin{table*}[h!]
\centering
\begin{tabular}{ll}
\bf Generated-Poe & \bf Generated-Marley   \\
\hline
her beautiful eyes were bright &   don t you know you ain t no fool\\
this day is a burden of tears & you r gonna make some fun \\
the darkness of the night & but she s fighting every rule \\
our dreams of hope and fears & ain t no sunshine when she s gone\\
\end{tabular}
\caption{\label{tab:ex}  Examples of the generated stylized quatrains. The punctuation is omitted since it was omitted in the training dataset.}
\end{table*}

\section{Experiments and evaluation}\label{ex}

The most standard approach for a comparison of two generative models would be to measure cross entropy loss at certain checkpoints. However, as \cite{Xie} writes: "There can be significant differences in final performance across checkpoints with similar validation losses." In our case cross entropy calculated in a straightforward manner does not give us any meaningful information. In order to quantitatively estimate our final model $G(C|S)$ we trained a plain vanilla LSTM without word-by-word document information support and with only classic word embeddings. We also trained a model with document information support but without bidirectional LSTMs for phonemes and characters included in the embeddings. All three models have shown comparable values of cross-entropy loss after an equal amount of epochs, which means that proposed additional structure is probably not facilitating learning but is likely not hindering it either.

\subsection{Sample cross entropy}
\label{ss:ce}

Cross entropy is one of the most natural theoretic-informational metrics to estimate the similarity of different texts. In order to distinguish this metric from the cross entropy loss, we call it the {\em sample cross entropy} and calculate it as described below. We sample several subsets with the same length (in words) from the original author texts in such a way that we end up with samples that contain a comparable number of unique texts for each author. We split the texts of a given author $A_i$ in two random groups and calculate the pairwise\footnotemark cross entropy between original texts of the author $A_i$ and texts generated by the model conditioned on that author $\{ T^{G}_i | A_i\}$.  The cross entropy between the sets of texts was calculated with MITML, see \cite{Hsu}, in the following manner: for every sample  written by the author and described above, we build a standard 3-gram based language model with standard MITML smoothing. We also build a common vocabulary across all samples. Then we calculate the perplexity by applying the language models based on the author-written texts to generative and original texts. After that, we apply logarithm to get the cross entropy instead of the perplexity, though both values in principle have a similar meaning. In Table \ref{tab:ce} one can see the results of these estimations. Analogous results for Russian can be found in Appendix in Table \ref{tab:ruce}. One can see that alongside with individual styles the model captures the {\em style of the time} mentioned earlier. Generated texts stylized for the authors from a similar time period tend to demonstrate lower sample cross entropies with human written texts written close to that time.

\addtocounter{footnote}{0}
\footnotetext{\noindent{Hence 'sample' in the name of the metric.}}

\begin{table*}[t!]
\centering
\begin{tabular}{llllllll}
\bf Model $G(A_i)/$ author & \bf Shakespeare & \bf Poe & \bf Carroll & \bf Wilde  & \bf Marley  & \bf Nirvana  & \bf MUSE   \\ \hline
\bf Generated-Shakespeare & \boldmath$19.0^{**}$ &  $21.6 $ & \boldmath$18.5^{*} $ & $19.9 $  &  $21.8  $ &  $22.0$  &  $22.4$  \\
\hline
\bf Generated-Poe & $22.0$  & \boldmath$20.4^{**} $ & $21.2$ & \boldmath$19.0^{*} $ &  $26.0 $  &  $25.4  $ &  $26.0$  \\
\hline
\bf Generated-Carroll & $22.2$  & $23.6$  & \boldmath$18.9^{*} $ & $22.5  $ & $22.4 $ & \boldmath$21.8^{**}$ & $23.8$  \\
\hline
\bf Generated-Wilde & $21.2$ & $20.9$ & \boldmath$20.5^{**}$  & \boldmath$18.4^{*} $ & $24.5$  & $24.8 $ & $26.4$ \\
\hline
\bf Generated-Marley & $24.1$  & $26.5 $ & $22.0  $& $27.0  $& \boldmath$15.5^{*}$  & \boldmath$15.7^{**}$   & $16.0$  \\
\hline
\bf Generated-Nirvana &  $23.7$ &  $26.2$ &  $20.0 $ &  $26.6  $&  $19.3  $& \boldmath$18.3^{*} $ & \boldmath$19.1^{**}$  \\
\hline
\bf Generated-MUSE &  $21.1$ &  $23.9$  &  $18.5$  &  $23.4 $ &  $17.4 $  & \boldmath$16.0^{**}$   & \boldmath$14.6^{*}$  \\
\hline
\hline
\bf Uniform Random &  $103.1$ &  $103.0$  &  $103.0$  &  $103.0 $ &  $103.5 $  & $103.3$   & $103.6$  \\
\hline
\bf Weighted Random &  $68.6$ &  $68.8$  &  $67.4$  &  $68.5 $ &  $68.5 $  & $68.0$   & $68.0$  \\
\hline
\hline
\hline
\bf SELF &  $23.4$ &  $21.8$  &  $25.1$  &  $27.3 $ &  $ 20.8 $  & $17.8$ & $13.3$  \\
\end{tabular}
\caption{\label{tab:ce}  Sample cross entropy between generated texts $\{ T^{G}_i | A_i\}$ and actual texts for different authors. The two smallest values in each row are marked with * and ** and a bold typeface. The sample cross entropy between random samples from the texts of the target author and randomly generated sequences of words (uniform and weighted respectively) as well as other samples written by the same author (denoted as SELF) are shown for reference.}
\end{table*}

\begin{table*}[t!]
\centering
\begin{tabular}{llllllll}
\bf Model $G(A_i)/$ author & \bf Pushkin & \bf Esenin & \bf Brodsky & \bf Letov  & \bf Zemfira    \\ \hline
\bf Generated-Pushkin & \boldmath$17.9^{*}$ &  \boldmath$21.8^{**}$ & $23.4$ & $27.0 $  &  $30.8 $ \\
\hline
\bf Generated-Esenin & \boldmath$20.4^{**}$  & \boldmath$18.8^{*} $ & $21.0$ & $22.7$ &  $26.0$  \\
\hline
\bf Generated-Brodsky & $23.5$  & \boldmath$21.1^{**}$  & \boldmath$17.2^{*} $ & $20.9$ & $23.8 $   \\
\hline
\bf Generated-Letov & $22.2$ & \boldmath${20.0^{**}}$ & $20.8$  & \boldmath$19.6^{*} $ & $23.6$ \\
\hline
\bf Generated-Zemfira &  $19.5$ &  \boldmath$17.1^{**}$ &  $18.1 $ &  $18.2$ &  \boldmath$16.6^{*} $ \\
\hline
\hline
\bf Uniform Random &  $103.0$ &  $103.1$ &  $103.0 $ &  $103.0$ &  $103.8$ \\
\hline
\bf Weighted Random &  $40.8$ &  $40.2$ &  $40.2 $ &  $42.6$ &  $45.6 $ \\
\hline
\hline
\hline
\bf SELF &  $35.0$ &  $33.7$  &  $38.0$  &  $28.3 $ &  $ 12.0 $  \\
\end{tabular}
\caption{\label{tab:ruce}  Sample cross entropy between generated texts $\{ T^{G}_i | A_i\}$ and actual texts for different authors. The two smallest values in each row are marked with * and ** and a bold typeface. The sample cross entropy between random samples from the texts of the target author and randomly generated sequences of words (uniform and weighted respectively) as well as other samples written by the same author (denoted as SELF) are shown for reference.}
\end{table*}

The lower is the sample cross entropy between the texts generated by the model and the texts written by every author the better the model captures author's writing style and vocabulary. The cross entropy between random samples from the texts of the same author demonstrates how {\em self-similar} the human-written texts are. Since an overwhelming amount of English text in our training dataset was text from the 20th century, the model 'perceives' texts of William Shakespeare or Edgar Allan Poe to be closer to the lyrics of Lewis Carrol and Oscar Wilde than to the samples of the original texts, however Shakespeare and Poe are also fairly well approximated by the model (it shows second best cross entropy there). To give a baseline we also provide cross-entropies between human-written texts and the texts sampled randomly out of the vocabulary as well as the the texts obtained through a weighted average sampling method. 

\subsection{BLEU}

Since BLEU is a metric estimating the correspondence between a machine's output and that of a human it is very natural to use it in order to measure the quality of the proposed model. For the experiments we sampled a random starting line out of the human-written poems and initialized the generative model with this line. Then we calculated BLEU between three actual lines that finished the human-written quatrain starting with a given first line and three lines generated by the model when initialized with the same human-written line. In Section \ref{formulation} we stated that one of the contributions of this paper is the idea to train the stylization model $G(C|S)$ on the whole corpus $C$ and then estimate the performance of $G(C|S)$ for different $S$. Table \ref{tab:bleu} illustrates this idea.

\begin{table}[b!]
\small
\begin{tabular}{llllllll}
\bf Model $G(A_i)$ & \bf Chosen author $S$ & \bf Validation dataset   \\ \hline
\bf  $G(S)$ & $33.0\%$ & $19.0\%$  \\
\bf $G(C|S)$ & $37.3\% (+13\%)$ & $37.6\% (+98\%)$  \\
\end{tabular}
\caption{\label{tab:bleu}  BLEU for the full model trained on one particular author dataset, $G(S)$, and on the whole dataset, $G(C|S)$, calculated on the chosen author validation dataset and on the validation dataset that includes a variety of authors. The results may vary across authors depending on the relative sizes of $S$ and $C$ but the general picture does not change.}
\end{table}

Indeed, not only the model $G(S)$ trained on texts of a particular author $S$ demonstrates the results that are worse than $G(C|S)$ when validated on the lyrics of the chosen author, $G(C|S)$ also performs almost two times better than $G(S)$ on the validation dataset containing texts from other authors.

Table \ref{tab:bleu2} shows BLEU calculated on the validation dataset for the plain vanilla LSTM, LSTM with author information support but without bidirectional LSTMs for phonemes and characters included in the embeddings and the full model. The uniform random and weighted random give baselines to compare the model to.

\begin{table}[b!]
\small
\begin{tabular}{llllllll}
\bf Model $G(A_i)$ & \bf BLEU   \\
\hline
\bf Uniform Random & $0.35\%$   \\
\bf Weighted Random & $24.7\%$   \\
\bf Vanilla LSTM & $29.0\%$   \\
\bf Author LSTM & $29.3\%$ ($+1\%$ to vanilla LSTM)    \\
\bf Full model  & $29.5\%$  ($+1.7\%$ to vanilla LSTM) \\
\end{tabular}
\caption{\label{tab:bleu2}  BLEU for uniform and weighted random random sampling, vanilla LSTM, LSTM with author embeddings but without phonetics, and for the full model. Phonetics is estimated to be almost as important for the task of stylization as the information on the target author.}
\end{table}

\subsection{Survey data}

\begin{table}[b!]
\centering
\begin{tabular}{llllll}
  & \bf Shak.& \bf Carroll & \bf Marley & \bf MUSE  & \bf LSTM  \\ \hline
\bf G.Shak. & \boldmath$0.37^{*}$ &  $0.04 $ & $0.05 $ & $0.14$  &  \boldmath$0.3^{*}$    \\
\bf R.Shak. & \boldmath$0.46^{*}$  & $0.05$ & $0.04$ & $0.07$ &   \boldmath$0.3^{*}$  \\ \hline
\bf G.Carroll & $0.02$  & $0.07$  & \boldmath$0.26^{*} $ & $0.18$ & \boldmath$0.41^{*}$  \\
\bf R.Carroll & $0.05$ & \boldmath$0.2^{*}$ & $0.14$  & $0.11$ & \boldmath$0.32^{*}$   \\ \hline
\bf G.Marley & $0.02$  & $0.01$ & \boldmath$0.47^{*}$& $0.2  $& \boldmath$0.29^{*}$   \\
\bf R.Marley &  $0.15$ &  $0.05$ &  \boldmath$0.4^{*}$ &  $0.1  $&  \boldmath$0.24^{*}$  \\\hline
\bf G.MUSE &  $0.09$ &  $0$  &  $0.12$  &  \boldmath$0.34^{*}$ &  \boldmath$0.39^{*} $    \\
\bf R.MUSE &  $0.03$ &  $0.05$  &  \boldmath$0.28^{*}$  &  \boldmath$0.39^{*} $ &  $0.2 $ 
 
\end{tabular}
\caption{\label{tab:te} Results of a survey with 140 respondents. Shares of each out of 5 different answers given by people when reading an exempt of a poetic text by the stylistic model of an author (prefaced with G. for {\em generated}) or by an actual author (prefaced with R. for {\em real}). The two biggest values in each row are marked with * and a bold typeface.}
\end{table}

\begin{table}[b!]
\begin{tabular}{llllll}
 &  \bf Pushkin & \bf Esenin & \bf Letov & \bf Zemf.  & \bf LSTM  \\ \hline
\bf G.Pushkin & \boldmath$0.31^{*}$ &  $0.22 $ & $0.02 $ & $0.0$  &  \boldmath$0.44^{*}$    \\
\bf R.Pushkin & \boldmath$0.62^{*}$  & $0.11$ & $0.03$ & $0.01$ &   \boldmath$0.23^{*}$  \\ \hline
\bf G.Esenin & $0.02$  & \boldmath$0.61^{*}$  & $0.08 $ & $0.0$ & \boldmath$0.29^{*}$  \\
\bf R.Esenin & $0.06$ & \boldmath$0.56^{*}$ & $0.07$  & $0.02$ & \boldmath$0.29^{*}$   \\ \hline
\bf G.Letov & $0.0$  & $0.02$ & \boldmath$0.40^{*}$& $0.08  $& \boldmath$0.51^{*}$   \\
\bf R.Letov &  $0.0$ &  $0.01$ &  \boldmath$0.61^{*}$ &  $0.02  $&  \boldmath$0.35^{*}$  \\\hline
\bf G.Zemfira &  $0.0$ &  $0.06$  &  $0.13$  &  \boldmath$0.4^{*}$ &  \boldmath$0.41^{*} $    \\
\bf R.Zemfira &  $0.0$ &  $0.02$  &  $0.08$  &  \boldmath$0.58^{*} $ &  \boldmath$0.31^{*}$  
\end{tabular}
\caption{\label{tab:ru} Results of a survey with 178 respondents. Shares of each out of 5 different answers given by people when reading an exempt of a poetic text by the stylistic model of an author (prefaced with G. for {\em generated}) or by an actual author (prefaced with R. for {\em real}). The two biggest values in each row are marked with * and a bold typeface.}
\end{table}

We randomly sampled 2 quatrains from William Shakespeare, Lewis Carroll, Bob Marley and MUSE band, and 2 quatrains generated by the model conditioned on those four authors respectively. Then 140 fluent English-speakers were asked to read all 16 quatrains in randomized order and choose one option out of five offered for each quatrain, i.e. the author of this verse is William Shakespeare, Lewis Carroll, Bob Marley, MUSE or an Artificial Neural Network. The summary of the obtained results is shown in Table \ref{tab:te}. Analogous results but for Russian language could be seen in Appendix in Table \ref{tab:ru} alongside with more detailed description of the methodology. It is important to note that the generated pieces for tests were human-filtered for mistakes, such as demonstrated in Table \ref{tab:ex}, whereas the automated metrics mentioned above were estimated on the whole sample of generated texts without any human-filtering.

Looking at Table \ref{tab:te} one can see the model has achieved good results in author stylization. Indeed the participants recognized Shakespeare more than 46\% of the times (almost 2.5 times more often than compared with a random choice) and did slightly worse in their recognition of Bob Marley (40\% of cases) and MUSE (39\% of cases, still 2 times higher than a random choice). This shows that the human-written quatrains were, indeed, recognizable and the participants were fluent enough in the target language to attribute given texts to the correct author.  At the same time, people were 'tricked' into believing that the text generated by the model was actually written by a target author in 37\% of cases for generated Shakespeare, 47\% for generated Marley, and 34\% for generated MUSE, respectively. Somehow, Lewis Carroll turned out to be less recognizable and was recognized in the survey only in 20\% of cases (corresponds to a purely random guess). The subjective underperformance of the model on this author can therefore be explained with the difficulty experienced by the participants in determining his authorship. 

\section{Conclusion}

{In this paper we have defined a problem of stylized text generation and have proposed an LSTM-based method for dealing with such tasks. We have also proposed a cross entropy based method to estimate the quality of stylization. The proposed LSTM is an extension of a language model which is supported by the document meta information at every step and works with large concatenated embeddings that include word embedding, a phoneme-based bidirectional LSTM final state, and a char-based bidirectional LSTM final state. We have successfully trained this model in Russian and in English. The texts generated by the model tend to be closer to the texts of the target author than the text generated by a plain vanilla LSTM both in terms of the cross sample entropy and BLEU. When faced with an author who is recognized by the participants of the test approximately two times more frequently than at random, participants mistakenly attribute the output of the proposed generative model to the target author as often as they correctly attribute original texts to the author in question. Such stylization can be of importance for more authentic dialogue interfaces and personalized human-machine interaction.}


\bibliographystyle{IEEEbib}
\bibliography{slt}

\end{document}